\documentclass[pdflatex,sn-mathphys-num,iicol]{sn-jnl}

\usepackage{graphicx}%
\usepackage{multirow}%
\usepackage{amsmath,amssymb,amsfonts}%
\usepackage{xcolor}%
\usepackage{textcomp}%
\usepackage{manyfoot}%
\usepackage{booktabs}%
\usepackage{algorithm}%
\usepackage{algorithmicx}%
\usepackage{algpseudocode}%
\usepackage{xspace}
\providecommand{\botrule}{\bottomrule}%

\newcommand{\mr}{multi-region\xspace}

\newcommand{\lite}{RETAS-lite\xspace}
\newcommand{\impact}{KB IMPACT\xspace}

\newif\ifbiofigmeasured
\newcommand{\biofigure}[1]{%
  \ifbiofigmeasured
    \global\biofigmeasuredfalse
    \includegraphics[width=25mm]{#1}%
  \else
    \global\biofigmeasuredtrue
    \includegraphics[width=25mm]{#1}\hspace{10pt}%
  \fi}

\hypersetup{pdftitle={Word-Level Alignment of OCR and Ground Truth under Zoning and Reading-Order Errors}}

\begin{document}

\title[Word Alignment of OCR Text Under Zoning Errors]{Word-Level Alignment of OCR and Ground Truth under Zoning and Reading-Order Errors}

\author*[1]{\fnm{Jorge Ram\'on} \sur{Fonseca Cacho}}\email{jorge.fonsecacacho@unlv.edu}

\affil*[1]{\orgdiv{Department of Computer Science}, \orgname{University of Nevada, Las Vegas}, \orgaddress{\city{Las Vegas}, \state{NV}, \country{USA}}. ORCID: 0000-0001-7917-5169}

\abstract{\paragraph*{Purpose}
OCR post-processing needs OCR text aligned word by word with its ground truth. Data sets rarely provide that mapping, and a monotone aligner can lose its place when zoning errors permute the reading order. We revisit this problem, the Zoning Problem of our 2019 paper, with a protocol that tests correctness under such permutations.

\paragraph*{Methods}
We apply OCR-side reading-order permutations with construction-time word-level gold and score correct-match F1, cascade length and recovery. We compare the 2019 location-aware aligner, a re-anchoring extension of it, a RETAS-style baseline (RETAS-lite), a multi-region aligner that combines unique-word and unique n-gram anchors with chunk-to-window assignment, genalog's RETAS and the original Java RETAS. Experiments use synthetic character noise with block moves, paragraph swaps and column interleaving, historical newspaper reading-order permutations, geometry-derived gold for Tesseract OCR, and a small word-box set.

\paragraph*{Results}
Synthetic tests show where local tracking fails after large displacements and where unique-anchor methods collapse when unique anchors are scarce. On our synthetic boilerplate page the 2019 method had the highest F1 among those tested. Historical newspaper and geometry-derived evaluations carry the limitations of approximate gold and the article-order convention. The first multi-region version has a quadratic scaling cost.

\paragraph*{Conclusion}
Local tracking remains fragile under large reading-order changes; the tested anchor-based methods handle many of those cases but fail when unique anchors are scarce. Approximate gold, the article-order convention, missing residual annotation, and the lack of modern multi-column word-level ground truth remain limitations.
}


\maketitle

\section{Introduction}\label{sec:intro}

Testing an OCR post-processing method~\cite{nguyen2021survey} on aligned word pairs requires knowing which ground-truth word, if any, corresponds to each OCR word. Data sets rarely come with this mapping. They usually give a page image, the OCR text, and a ground truth text, and the mapping has to be recovered by aligning the two texts. Alignment errors then become label errors: a corrector trained on misaligned pairs learns to turn one word into an unrelated one, and an evaluation run on misaligned pairs reports the wrong error rate~\cite{nguyen2021survey,chiron2017icdar,rigaud2019icdar}.

For short or clean texts, established sequence-alignment methods are available. Dynamic programming finds an optimal alignment under edit distance~\cite{needleman1970general,wagner1974string}, and anchor-based schemes such as RETAS make it fast enough for whole books~\cite{yalniz2011fast}. The hard case is zoning. When an OCR engine segments a page into the wrong regions, or reads the regions in the wrong order, the OCR text is a permutation of the ground truth at the block level, on top of the usual character errors~\cite{kanai1995automated}. A monotone alignment cannot represent that permutation, and a method that follows its position in the ground truth can lose its place and fail for many words in a row.

In 2019 we described this as the Zoning Problem~\cite{fonsecacacho2019aligning}. Our location-aware alignment algorithm, tested on the TREC-5 Confusion Track~\cite{kantor2000trec5}, reached an average alignment accuracy of 98.547\% on documents without zoning problems and 81.07\% on documents with them. We proposed ways to close that gap but left them as future work. This paper picks up from there. Seven years on, the question is still practical: digitized newspapers and other multi-column collections are exactly where reading-order errors are common, and they are also where many post-OCR training sets come from.

We started by re-testing the problem, and the first lesson was about measurement. An early version of our test harness moved an out-of-order block in both the OCR text and the ground truth, so there was no reading-order error to recover from, and it counted every word that received some match as a success, whether or not the match was correct. Under that harness every method looked perfect. We rebuilt the harness so that the reading order changes on the OCR side only, every OCR word has a known gold position, and we count correct matches, the length of each error cascade, and how quickly a method recovers. The old ``matched'' rate is still reported where it helps comparison with 2019, but never as a correctness score.

With the synthetic measurement corrected, we compare six aligners (counting the two frozen versions of the multi-region aligner as one) on synthetic zoning errors, on real newspaper reading order, on real Tesseract OCR~\cite{smith2007overview} of historical newspapers from the \impact collection~\cite{papadopoulos2013impact,kbimpact}, on a small set of pages with human word boxes, and on TREC-5. The contributions are:
\begin{itemize}
\item A measurement protocol for alignment under zoning errors: OCR-side reading-order permutation with word-level gold, precision, recall and F1 of correct matches, cascade length and recovery rate, with bootstrap confidence intervals and paired tests (Sect.~\ref{sec:measure}).
\item A gold alignment for real OCR derived from page geometry, which matches OCR word boxes to ground-truth word boxes by overlap and edit distance, with an explicit high-confidence subset and an error estimate (Sect.~\ref{sec:geogold}).
\item A multi-region aligner that combines unique-word and unique n-gram anchors with greedy assignment of OCR chunks to ground-truth windows. It is simple and its parameters were frozen before the real-OCR tests (Sect.~\ref{sec:mr}). The comparison includes our RETAS-style baseline, genalog's RETAS~\cite{gupte2021lights,genalog2021software} and the original Java RETAS~\cite{retas2013software} (Sect.~\ref{sec:res-geo}).
\item An account of where it fails: the first version has quadratic scaling and becomes slower than the baseline somewhere between 10{,}000 and 25{,}000 tokens, and like the other anchor-based methods tested it collapses on our synthetic boilerplate page with few distinct word types, where our 2019 location-aware method has the highest F1 among the tested methods on that fixture (Sects.~\ref{sec:res-stress} and~\ref{sec:res-long}). A second frozen version reduces the length problem.
\end{itemize}
We also report what we could not reproduce. TREC-5 does not label which documents have zoning problems, so the 2019 split into 98.5\% and 81\% cannot be rebuilt; we report TREC-5 results as a reproduction attempt only (Sect.~\ref{sec:res-trec}).

Our comparison covers a limited set of implementations and settings. We do not infer a field-wide ranking from it.

\section{Related work}\label{sec:related}

\paragraph{Sequence alignment.} The classic tools for aligning two strings are dynamic programming over edit operations, introduced for protein sequences by Needleman and Wunsch~\cite{needleman1970general} and for string correction by Wagner and Fischer~\cite{wagner1974string}. Both run in time proportional to the product of the two lengths, which can be prohibitively slow for whole books, and both produce a monotone alignment: if word $i$ is aligned before word $j$ in one text, it is aligned before it in the other. A reading-order error breaks that assumption, so a monotone alignment cannot preserve all correct correspondences between a displaced block and the text whose order it crosses.

\paragraph{Anchor-based alignment.} Heckel~\cite{heckel1978technique} observed that lines occurring exactly once in each of two files can serve as candidate anchors, and that the regions between anchors can then be compared separately. The same idea, at the word level, underlies the main OCR alignment methods for long texts. Feng and Manmatha aligned OCR output with book text through a hierarchical scheme based on hidden Markov models~\cite{feng2006hierarchical}. Yalniz and Manmatha's Recursive Text Alignment Scheme (RETAS)~\cite{yalniz2011fast} aligns words that are unique in both texts, recurses into the segments between matched unique words until the segments are small, and finishes with edit-distance alignment. The original Java implementation is available~\cite{retas2013software}, and Microsoft's genalog package~\cite{gupte2021lights,genalog2021software} includes a Python reimplementation that uses Needleman--Wunsch on the final segments. RETAS was designed for books, where the order of the text is reliable, and it treats the text between two consecutive matched anchors as one segment.

\paragraph{Location-aware alignment.} Our 2019 algorithm~\cite{fonsecacacho2019aligning} takes a different route. It walks through the OCR text one word at a time, keeps a pointer into the ground truth, and looks for each OCR word in a small window around that pointer, first exactly and then with increasing edit distance. It does not need unique words, and it is fast, but a block that moves farther than the window sends the pointer astray. The 2019 paper reported this as the Zoning Problem and measured its cost on TREC-5.

\paragraph{Zoning and reading order.} Kanai et al.~\cite{kanai1995automated} measured zoning quality by the number of edit operations, including block moves, needed to turn OCR output into the correct text. Clausner, Pletschacher and Antonacopoulos~\cite{clausner2020flexible} proposed a character accuracy measure that does not depend on reading order. Both evaluate OCR output; neither produces a word-level alignment that can be used as training data. Ground truth formats such as PAGE~\cite{pletschacher2010page} store the reading order explicitly, which is what makes our reading-order and geometry-gold tests possible.

\paragraph{Data sets.} The TREC-5 Confusion Track~\cite{kantor2000trec5,trec5data} provides 1994 Federal Register documents in three versions: the original typeset text and two OCR versions with roughly 5\% and 20\% character error. The IMPACT project produced ground truth for historical documents in PAGE format~\cite{papadopoulos2013impact}; the Dutch national library (KB) released its IMPACT newspaper ground truth into the public domain~\cite{kbimpact}. Post-OCR competitions at ICDAR~\cite{chiron2017icdar,rigaud2019icdar} and recent surveys~\cite{nguyen2021survey} show how much of the field depends on aligned OCR and ground truth pairs.

\paragraph{Non-monotone reference collation.} ACDC~\cite{smith2023acdc} aligns noisy HTR with digital editions or other witnesses using a character-level HMM whose reference read head can jump backwards or forwards by an arbitrary distance. Character-subsequence pruning limits the search; the model is implemented as \texttt{seriatim} in passim v2. ACDC selects aligned lines for recognizer retraining. Non-monotone reference alignment and alignment-derived training supervision are therefore established prior work, not general novelty claims of this paper. Our study instead examines the word-link protocol of Sect.~\ref{sec:measure} and frozen implementations; ACDC has not been evaluated in our comparison.

\paragraph{Text reuse and multi-token links.} Miller, Kuflik and Lavee~\cite{miller2025textalignment} combine Smith--Waterman alignment with edit/fastText similarity, word-boundary split/merge handling, multi-token traceback and post-alignment transposition matching. They evaluate aligned token-position tuples, including many-to-one links, against expert Hebrew/Aramaic alignments. Thus neither position-link evaluation nor split/merge-aware alignment is new in general. Their text-reuse task also permits semantic, acronym and morphological correspondences, whereas our present single-index word mapping targets literal OCR occurrences; the scores are not interchangeable.

\paragraph{Dutch parallel-corpus construction.} Debaene et al.~\cite{debaene2025evaluating} compare RETAS with sentence-embedding retrieval within each early modern Dutch play and with a newline-aware extension of that retrieval that generates concatenations of consecutive segments. Their triadic construction chooses the lowest-CER candidate across the three approaches for each ground-truth sentence. This is directly relevant to OCR/reference corpus construction in Dutch, but sentence-level CER selection is a different criterion from independently verified word-occurrence links. We have not tested their retrieval methods here.

\paragraph{Granularity-adaptive evaluator matching.} MinerU2.5-Pro~\cite{wang2026minerupro} introduces Multi-Granularity Adaptive Matching (MGAM) for document-parsing evaluation. Their Section~5.2 keeps ground-truth segmentation fixed and compares Hungarian matching at the original prediction granularity, after line-break splitting, and after consecutive split/merge partition enumeration, choosing the best aggregate candidate. Text elements use edit similarity, and predicted tables can be converted to text for matching. This is prior art for adaptive split/merge assignment, but its evaluator element matches are not independent word-occurrence gold. We do not infer comparative alignment accuracy from its parsing scores.

\paragraph{Reference-grounded line annotation.} RefLAM~\cite{guechaoui2026reflam} combines page-signature anchoring with greedy fuzzy line-to-reference-window matching, separate main/margin handling and human review of every included line. Its confidence-100 rule establishes normalized-string identity, not a unique occurrence or a pixel-perfect box (their Remark~1). It is a close pointer/window reference-grounding pipeline, but its line-level review task differs from our single-index word-link evaluation. We do not treat its annotation throughput as an aligner speed comparison.

\section{Measuring alignment under zoning errors}\label{sec:measure}

\subsection{Task}

Let $G = g_1 \dots g_m$ be the ground truth words of a page and $O = o_1 \dots o_n$ the OCR words. An aligner returns a mapping $a: \{1..n\} \to \{1..m\} \cup \{\varnothing\}$ that assigns each OCR word a ground-truth position or leaves it unmatched. The evaluation interface records at most one ground-truth position per OCR word; this does not itself guarantee an injective mapping across OCR words. In the gold alignment, merged and split words are left unmatched or represented by one of their parts. A gold alignment $a^*$ gives the correct position, or $\varnothing$ for OCR words with no source.

\subsection{Metrics}

We count a true positive when $a(i) = a^*(i) \neq \varnothing$. When the gold position is nonempty, a different nonempty prediction counts as both a false positive and a false negative, and an unmatched prediction counts as a false negative. When the gold position is $\varnothing$, any match counts as a false positive. Precision, recall and F1 follow as usual. This differs from the success rate used in 2019 and in our first test harness, which counted the share of OCR words that received any match. We keep that share under the name \emph{matched} where it helps comparison with earlier numbers, but it is not a measure of correctness: an aligner that matches every word to the wrong position scores 100\% on it.

Zoning errors produce cascades. Let a \emph{break} be an OCR position $i>1$ for which both $a^*(i)$ and $a^*(i-1)$ are nonempty and $a^*(i) \neq a^*(i-1)+1$. The first position and transitions involving an OCR word with gold $\varnothing$ are excluded. The cascade length at a break is the number of consecutive incorrectly aligned OCR words starting at the break, and $\bar{C}$ is its mean over breaks. The recovery rate $R_N$ is the share of breaks followed by at least one correct alignment among the next $N$ OCR words; we report $N = 10$.

Confidence intervals are 95\% percentile bootstrap intervals over seeds (synthetic data) or pages (real data), with 2{,}000 resamples~\cite{efron1994bootstrap}. To compare two aligners on the same items we use a paired bootstrap with 5{,}000 resamples of the per-item difference and report its mean, interval, and $p$, twice the share of bootstrap means on the other side of zero (including zero), capped at one. A reported $p < 0.001$ means that none of the 5{,}000 bootstrap means fell on the opposite side of zero. These intervals describe the evaluated seeds or pages; they are not claims about a larger population.

\subsection{A generator with known gold}\label{sec:generator}

The synthetic test starts from a fixed ground truth of 517 words of mostly Federal-Register-style prose with 327 distinct word types (case-folded). The OCR side is a permuted copy, and the gold alignment is the permutation itself, so every OCR word keeps the index of the ground-truth word it came from. We use three zoning patterns:
\begin{itemize}
\item \emph{Block move}: a contiguous block (4\%, 8\% or 15\% of the words, and at least eight words long) is cut from the first half of the page and inserted near the end, well beyond a local search window.
\item \emph{Paragraph swap}: the page is split into paragraphs of $\max(20, \lfloor n/(2k+2) \rfloor)$ words, where $n$ is the page length and $k$ the number of swap steps (86 words at medium severity), and the paragraphs are reordered by a permutation that is guaranteed not to be the identity (1, 2 or 3 swap steps).
\item \emph{Column interleave}: the ground truth is read as a left column followed by a right column, and the OCR side alternates chunks of 8, 12 or 16 words from the two columns, as when lines of two columns are read across.
\end{itemize}
The three numbers in each pattern are the mild, medium and severe settings. For block moves a higher setting displaces more text; for paragraph swaps it adds swap steps over shorter paragraphs; for column interleave it only makes the chunks longer, which means fewer breaks, so ``severe'' interleave is not necessarily harder than ``medium''. The chunk lengths 8, 12 and 16 are also the chunk sizes of the multi-region aligner (Sect.~\ref{sec:mr}), so its column-interleave results may be helped by this match; the real-OCR tests do not have this property. After permuting, each word longer than two characters receives, with probability $p$, one random character substitution, insertion or deletion; $p$ is 2\%, 6\% or 10\% in the main sweep and up to 25\% in the stress tests. The noisy word keeps its gold position. Unless stated otherwise we report medium severity with $p = 6\%$ over seeds 0--9.

The base text was written to contain no repeated paragraphs. An earlier version built a longer page by repeating the base text six times; every column half then had an exact copy elsewhere on the page, and on paragraph swap every aligner locked onto the sequential copy and produced the same mapping. We report only results from the non-repeating generator.

\section{Aligners}\label{sec:methods}

All aligners take the two word sequences, lowercased and split into runs of word characters, and return at most one ground-truth index per OCR token; global injectivity is not assumed for every baseline. Edit distance is character-level Levenshtein distance between words~\cite{wagner1974string}.

\subsection{Location-aware alignment (2019)}\label{sec:loc}

We reimplemented the 2019 algorithm~\cite{fonsecacacho2019aligning} from its published description. It keeps a pointer $p$ to the last matched ground-truth position. For each OCR word in order it tries, and stops at the first success: an exact match at $p+1$; an exact match in the forward window $[p+1, p+w]$; an exact match in the backward window $[p-w, p-1]$; and then, for edit distance $d = 1, 2, \dots, d_{\max}$, the first word within distance $d$ in the forward window and then in the backward window. A match moves the pointer to the matched position. A failure leaves the word unmatched and advances the pointer by one, on the assumption that the OCR word was a garbled version of the next ground-truth word. The search uses $w = 20$ and $d_{\max} = 3$ and tries the forward window before the backward window, matching the 2019 settings. Where the published description is silent on whether confidence weights affect match selection, we record the weights, for example for dictionary words, but they do not change which match is chosen.

The weakness is visible in the description. When a block of the OCR text comes from a place farther than $w$ words away, the correct positions can lie outside the search windows. Each failure advances the pointer by one word, which can cause drift and further failures in the text that follows.

\subsection{Re-anchoring}\label{sec:reanchor}

The re-anchoring extension was our first attempt at the open problem. It runs the 2019 algorithm, then scans the result for streaks of at least four consecutive unmatched OCR words. For each streak it searches for a short sequence, four words and then three, that starts within the first 80 positions of the streak and agrees with some ground-truth sequence from 80 positions before to 200 positions after the last good match, in all but at most one word, where a word agrees if it is equal or within edit distance 1. If one is found, the next 40 OCR words are matched greedily forward from the anchor within the next 60 ground-truth positions, overwriting earlier matches. The search is still local to the last good match, so it repairs drift after short displacements but cannot find a block that came from far away.

\subsection{RETAS-lite}\label{sec:lite}

As a fast stand-in for RETAS we use a single-level version that we call \lite. It takes as anchors all words that occur exactly once in the OCR text and exactly once in the ground truth, sorted by OCR position, and matches each anchor to its ground-truth twin. Between two consecutive anchors it matches each OCR word to the closest ground-truth word within edit distance 2, searching forward from the last match for at most 40 positions and not past the next anchor. Unlike RETAS it does not recurse, and it does not discard anchors that are out of order; the second difference is what lets it follow a moved block, because the unique words inside the block still point to the right place. \lite is our own reimplementation and should not be read as a measurement of RETAS itself; for that we run genalog and the original Java tool (Sect.~\ref{sec:external}).

\subsection{Multi-region alignment}\label{sec:mr}

The \mr aligner keeps the anchor idea of \lite and adds a second stage that is built for out-of-order material. It has four steps (Algorithm~\ref{alg:mr}).

\emph{1. Anchors.} Words that are unique on both sides are anchors, as in \lite. Then, for $k = 4, 3, 2$ in that order, every $k$-gram of words that occurs exactly once in the OCR text and exactly once in the ground truth, and does not overlap positions already anchored on either side, adds $k$ anchors. The $k$-grams find anchors in text whose individual words are common.

\emph{2. Local fill.} The anchors are sorted by OCR position. For each consecutive pair whose ground-truth positions are also increasing, the unmatched OCR words between them are matched to the closest unused ground-truth word between the two anchors, within edit distance $e = 2$ and at most $L = 60$ positions ahead of the last match.

\emph{3. Chunk-to-window assignment.} This step handles material that the anchors left unmatched, which after a zoning error is mostly the displaced text. For chunk sizes $s = 16, 12, 8$ in turn, the OCR text is cut into consecutive chunks of $s$ words, and chunks with at least $\max(2, \lfloor s/3 \rfloor)$ unmatched words are kept. The ground truth is cut into windows of $s$ words with a stride of $s/2$, and windows that contain any matched position are dropped. Each chunk--window pair is scored by the size of the multiset intersection of their words, and pairs with a score below 2 are dropped. The pairs are accepted greedily from the highest score down, each chunk at most once and no two accepted windows overlapping, and each accepted pair is filled locally as in step 2. Because the windows can be anywhere in the ground truth, a chunk can be placed far from its neighbours, which is what a reading-order error requires.

\emph{4. Final sweep.} Each OCR word still unmatched is matched to the first unused exact twin in the ground truth, searching forward from the position after the previous matched word and wrapping around, and otherwise to any unused word within edit distance 1.

\begin{algorithm}[t]
\caption{Multi-region alignment (v1). $O$: OCR words, $G$: ground truth words.}\label{alg:mr}
\begin{algorithmic}[1]
\State $A \gets$ unique-word anchors $\cup$ non-overlapping unique $k$-gram anchors, $k = 4, 3, 2$
\State $a(i) \gets j$ for every anchor $(i, j) \in A$
\For{consecutive anchors $(i_1, j_1), (i_2, j_2)$ with $j_1 < j_2$}
  \State \Call{Fill}{$O[i_1{+}1..i_2{-}1]$, $G[j_1{+}1..j_2{-}1]$} \Comment{edit distance $\le 2$, look-ahead $\le 60$}
\EndFor
\For{$s \in (16, 12, 8)$}
  \State $\mathcal{C} \gets$ OCR chunks of $s$ words with $\ge \max(2, \lfloor s/3 \rfloor)$ unmatched words
  \State $\mathcal{W} \gets$ GT windows of $s$ words, stride $s/2$, with no matched position
  \State $P \gets \{(c, w, |c \cap w|) : c \in \mathcal{C}, w \in \mathcal{W}, |c \cap w| \ge 2\}$, sorted by score
  \For{$(c, w, \cdot) \in P$ with $c$ unassigned and $w$ disjoint from assigned windows}
    \State assign $c \to w$; \Call{Fill}{$c$, $w$}
  \EndFor
\EndFor
\State match leftovers to the next unused exact twin, else to any unused word within distance 1
\end{algorithmic}
\end{algorithm}

The parameters are the anchor sizes $(4, 3, 2)$, chunk sizes $(16, 12, 8)$, $e = 2$, $L = 60$ and the minimum chunk score of 2. They were set during the synthetic and reading-order experiments of Sects.~\ref{sec:res-synth} and~\ref{sec:res-ro} and then frozen, under the tag \texttt{multiregion-v1-frozen}, before any of the stress, real-OCR, word-box or long-document tests. The v1 code freeze is \texttt{e7813fc}; the v2 code freeze is \texttt{83bb593}. The tag commits \texttt{9fa4083} and \texttt{ca17f76} record those freezes rather than changing the algorithms. No value was changed afterwards; every later change went into a separate version. The results of Sects.~\ref{sec:res-synth} and~\ref{sec:res-ro} are therefore in-sample for \mr v1, and only the later tests are held out.

The aligner is not new in its parts. Unique-element anchors go back to Heckel~\cite{heckel1978technique} and RETAS~\cite{yalniz2011fast}, and window matching by bag-of-words overlap is a common heuristic. We study this particular ordering of the steps and unconstrained chunk placement, rather than claiming non-monotone reference alignment as a new idea; ACDC already permits arbitrary reference jumps~\cite{smith2023acdc}. Step~3 compares every chunk with every window, so its cost grows with the product of the two lengths; we return to this in Sect.~\ref{sec:res-long}.

\subsection{Multi-region v2}\label{sec:v2}

After the long-document and boilerplate results of Sect.~\ref{sec:results} we built a second version with three changes, developed only on synthetic seeds 100--129 and on IMPACT pages outside the 200-page test set, then frozen as \texttt{multiregion-v2-frozen} and evaluated once on the held-out data:
\begin{itemize}
\item \emph{Indexing.} Chunk-to-window scoring uses an inverted index from words to candidate windows and precomputed word counts per window. On a development document of 10{,}413 tokens, with the long-document fallback left off, this was 3.9 times faster than v1 at the same F1.
\item \emph{Long-document fallback.} Documents of 8{,}000 OCR tokens or more are handed to \lite. On development data this beat both no fallback and a windowed variant in accuracy and runtime at 10{,}000 and 25{,}000 tokens.
\item \emph{Scarcity merge.} If fewer than 5\% of tokens are unique on both sides and fewer than 2\% of word 3-grams are, or if fewer than 35\% of OCR words are matched after the v1 steps, the unmatched spans are aligned with the 2019 location-aware algorithm and merged in. The idea is to use the 2019 positional prior where anchors are scarce.
\end{itemize}
We wrote a success criterion into the proposal before the held-out run: on boilerplate pages, v2 should beat v1 by at least 0.15 F1 with paired $p < 0.01$, without losing more than 0.01 on column interleave. Sect.~\ref{sec:res-long} shows that the first part was not met.

\subsection{External baselines}\label{sec:external}

\emph{genalog.} We use the RETAS reimplementation in Microsoft's genalog (MIT licence)~\cite{genalog2021software}, \texttt{align\_w\_anchor}, which aligns characters and returns, for each OCR word, the ground-truth words it overlaps. When there are several we keep the one with the smallest edit distance to the OCR word.

\emph{Java RETAS.} We compiled the original Recursive Text Alignment Scheme (version 1.1, GPL-3.0)~\cite{retas2013software} from the Early Modern OCR Project mirror with JDK~17, ran it at word level, and parsed its aligned output into a word mapping. On one document pair, the tool's reported accuracy (0.495) and the F1 of our parsed mapping (0.493) agreed to within 0.002, a check of the parsing rather than a general proof of it.

\section{Data and protocol}\label{sec:setup}

\subsection{Data}

\paragraph{Synthetic zoning.} We use the generator of Sect.~\ref{sec:generator} with seeds 0--9. For the stress tests we vary the noise rate from 10 to 25\% under column interleave, split the page into 8, 15 or 30 regions, use short pages of about 96 words with a paragraph swap, and use a boilerplate page of 693 words containing 59 distinct word types (only 10 occur once), modelled on the repetitive notices of the Federal Register, with medium column interleave and 6\% noise. For long documents we concatenate distinct paragraphs, with no repetition, to 1{,}000--50{,}000 tokens and apply medium column interleave with 6\% noise (five seeds up to 10{,}000 tokens, three above).

\paragraph{KB IMPACT reading order.} The KB IMPACT newspaper ground truth~\cite{kbimpact} contains 1{,}024 newspaper pages from 1618 to 1885 in PAGE format~\cite{pletschacher2010page}, with text regions and PAGE reading-order groups. Ordering between articles inside a top-level UnorderedGroup is not specified by that annotation. The KB released it as public domain (CC0). For a first real test without OCR noise the historical fixture takes as $G$ the ground-truth text in the XML traversal order of region references (a linearization convention, not a complete annotated page order) and the same text in geometric order (top to bottom, using column bins when the page has columns) as $O$; the gold alignment is the index map within each region. We shuffled the sorted page list with seed 0. Of the first 50 pages we checked, 40 have a geometric order that differs from the XML traversal order, and we use those 40.

\paragraph{KB IMPACT with real OCR.} \label{sec:geogold} For the main test we ran Tesseract~5~\cite{smith2007overview} (\texttt{nld+eng}, page segmentation mode 6) on the KB page images of 200 pages drawn with seed 42, and used at most the first 600 OCR words of each page. Five of those pages also appear in the 40-page reading-order fixture; the frozen parameters were not chosen on either set. There is no word-level alignment between Tesseract output and the ground truth, so we derived one from geometry. The PAGE files give boxes for text regions only, so we split each region box into line strips and each line into word boxes in proportion to character counts. Each Tesseract word box is then matched greedily, one to one, to the ground-truth word box that maximizes $\mathrm{IoU} - 0.75\,\mathrm{NED}$, where NED is the edit distance normalized by the longer word; a match is accepted if this score is positive and either IoU $\ge 0.05$ or NED $\le 0.2$, with a fallback to the nearest word in the same region when NED $\le 0.2$. A gold match is \emph{high-confidence} if the words are equal, or if NED $\le 0.15$ and IoU $\ge 0.05$.

This gold is approximate. We automatically classified every matched pair on 20 pages as exact, close (NED $\le 0.25$) or mismatched, and estimated its error as the share of mismatches plus a quarter of the close pairs. The estimate was 0.238, with 62.6\% of matched pairs exact and 64.7\% of OCR words covered. Summed over the same 20 pages the stored type counts are 17{,}894 exact matches, 3{,}845 close (soft-edit) matches, 3{,}452 text mismatches and 14{,}882 unmatched OCR tokens (25{,}191 matched pairs among 40{,}073 OCR tokens). That quality check uses every OCR token on those 20 pages, not the 600-word cap used in the alignment evaluation. These are micro totals; the 0.238 error estimate and the 62.6\% and 64.7\% figures above remain page-mean summaries and are not the corresponding micro shares. The estimate counts disagreements between the OCR and ground-truth spellings of box-matched words, so it mixes gold errors with OCR errors, and it does not see wrong matches between similar words; it should be read as a rough indication rather than a bound. For these reasons we report F1 on all gold matches and on the high-confidence subset, with the approximate-gold limitations stated above.

\paragraph{Word-box gold.} To check the geometry gold where real word boxes exist, we used three CC0 corpora of 16th- and 17th-century prints with human word coordinates in PAGE format (17\_frak\_simple, 17\_frak\_complex and 16\_ant\_simple~\cite{boenig17fraksimple}), 14 pages in total, OCRed with Tesseract (\texttt{deu+eng}) and matched by the same IoU and NED rule against the true word boxes. Here the gold uses human-annotated word boxes, but the pages are blackletter and antiqua prints on which Tesseract is weak, and the same error estimate is 0.447 with 64.9\% coverage, which may reflect OCR errors as well as matching errors. We found no open data set of modern multi-column pages with human word boxes under a CC0, CC-BY or public-domain licence.

\paragraph{TREC-5.} The Confusion Track~\cite{kantor2000trec5,trec5data} is the data of the 2019 paper. We matched 479 original documents with their degrade5 versions by file name and sampled 30 documents (seed 0), using at most the first 600 tokens of each for the frozen comparison. The track has no word-level alignment and no zoning labels. We therefore score agreement with genalog's alignment as a soft reference and the matched rate, and we also build a synthetic-gold variant: 20 original documents (first 500 tokens) with medium column interleave and 10\% noise from our generator. The TREC-5 tarball is served from the NIST track page without an agreement form; we used it for research only and do not redistribute it.

\subsection{Protocol}

The \mr v1 parameters were frozen before the stress tests, and v2 was developed on separate seeds and pages and evaluated once. The 200 IMPACT test pages were held out from parameter development. Their geometry gold was not inspected while v2 was built, apart from the automatic 20-page quality classification described above, which uses a subset of these 200 pages. All aligners were run on an AMD EPYC 9175F CPU using Python 3.12, one document at a time; runtimes are wall-clock seconds per document and include only the alignment call. Synthetic results use seeds 0--9 unless noted; real-data intervals are over pages.

Use of AI assistants in preparing the manuscript and supporting code is documented in the Statements and Declarations.

\section{Results}\label{sec:results}

\subsection{Synthetic zoning errors}\label{sec:res-synth}

\begin{table*}[t]
\caption{Synthetic zoning ablation: word-alignment F1 against gold, mean over 10 seeds with 95\% bootstrap CI (medium severity, 6\% character noise, 517-token base text). Reading order is permuted on the OCR side only.}\label{tab:synth}
\centering\small
\begin{tabular}{@{}lccc@{}}
\toprule
Aligner & Block move & Paragraph swap & Column interleave \\
\midrule
Location-aware (2019) & 0.890 [0.865, 0.913] & 0.012 [0.003, 0.029] & 0.255 [0.242, 0.265] \\
+ re-anchoring & 0.922 [0.907, 0.935] & 0.339 [0.338, 0.340] & 0.353 [0.348, 0.358] \\
RETAS-lite & 0.998 [0.996, 0.999] & 0.977 [0.975, 0.977] & 0.907 [0.904, 0.909] \\
genalog (RETAS) & 0.920 [0.920, 0.921] & 0.181 [0.164, 0.215] & 0.500 [0.499, 0.501] \\
Multi-region v1 & 1.000 [0.999, 1.000] & 1.000 [1.000, 1.000] & 0.983 [0.981, 0.985] \\
\botrule
\end{tabular}
\end{table*}

\begin{table*}[t]
\caption{Cascade behaviour on the synthetic ablation of Table~\ref{tab:synth}: mean cascade length $\bar{C}$ (consecutive wrong words after a zoning break) and recovery rate $R_{10}$ (share of breaks followed by a correct match within 10 words). Means over 10 seeds.}\label{tab:cascade}
\centering\small
\begin{tabular}{@{}lcccccc@{}}
\toprule
 & \multicolumn{2}{c}{Block move} & \multicolumn{2}{c}{Paragraph swap} & \multicolumn{2}{c}{Column interleave} \\
\cmidrule(lr){2-3}\cmidrule(lr){4-5}\cmidrule(lr){6-7}
Aligner & $\bar{C}$ & $R_{10}$ & $\bar{C}$ & $R_{10}$ & $\bar{C}$ & $R_{10}$ \\
\midrule
Location-aware (2019) & 25.3 & 0.33 & 129.0 & 0.17 & 48.8 & 0.23 \\
+ re-anchoring & 18.2 & 0.53 & 66.5 & 0.17 & 23.6 & 0.33 \\
RETAS-lite & 0.7 & 1.00 & 3.3 & 0.98 & 1.3 & 1.00 \\
genalog (RETAS) & 13.7 & 0.67 & 207.0 & 0.07 & 6.3 & 0.51 \\
Multi-region v1 & 0.0 & 1.00 & 0.0 & 1.00 & 0.1 & 1.00 \\
\botrule
\end{tabular}
\end{table*}

Table~\ref{tab:synth} shows the ablation on the synthetic generator. The 2019 location-aware algorithm handles a single moved block reasonably well (F1 0.890), because the moved block and the text right after it are lost, but it fails on paragraph swaps (0.012) and on interleaved columns (0.255). Table~\ref{tab:cascade} shows why: after a paragraph swap its cascade runs for 129 words on average, longer than one 86-word paragraph, and only one break in six is followed by a correct match within ten words. This is the Zoning Problem of 2019 in its cleanest form.

Re-anchoring helps, but not enough to recover from large reorderings. It raises F1 to 0.922, 0.339 and 0.353 and roughly halves the cascades on paragraph swap and column interleave. We checked that the gain comes from re-anchoring itself and not from noise in the harness: on seed 0 of a medium paragraph swap with 2\% noise it fired nine times and changed 145 mappings, raising F1 from 0.003 to 0.338. Its search, however, stays near the last good match, and a paragraph that moved by a third of the page is out of reach.

The anchor-based methods are in a different range. \lite scores 0.998, 0.977 and 0.907, with mean cascades of 0.7 to 3.3 words, which confirms that unique words are good anchors even when the text is out of order, as long as the aligner does not insist on keeping the anchors in order. The \mr aligner reaches 1.000, 1.000 and 0.983, and every break is followed by a correct match within ten words. genalog's RETAS scores 0.920 on block moves but 0.181 on paragraph swaps and 0.500 on column interleave. It is also two to three orders of magnitude slower on these 517-word pages, 0.55--0.81\,s against 0.001--0.002\,s for \lite and \mr.

\begin{figure*}[t]
\centering
\includegraphics[width=\linewidth]{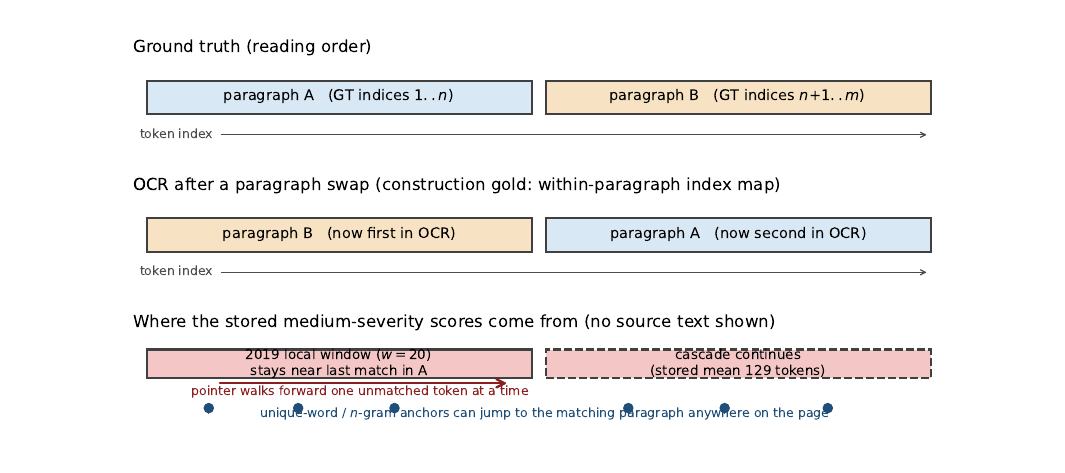}
\caption{Token-index schematic of an OCR-side paragraph swap. Blocks are labelled by ground-truth index ranges, not by source text. Construction gold maps each OCR token to its original index. The 2019 tracker keeps a local window near the last accepted match ($w=20$), so after the swap it continues searching near paragraph~A while the OCR stream reads paragraph~B; the stored mean cascade on the medium paragraph-swap condition is 129 tokens (Table~\ref{tab:cascade}). Unique-word and $n$-gram anchors are not bound to that window.}\label{fig:qualzoning}
\end{figure*}

Figure~\ref{fig:qualzoning} is a token-index schematic of that paragraph-swap case. We draw blocks by ground-truth index rather than source text, so the figure does not quote corpus wording. After the swap, the first OCR paragraph should map to paragraph~B and the second to paragraph~A. The 2019 tracker keeps a local window of $w=20$ near the last accepted match. It is then still looking near the start of~A while the OCR stream is reading~B, and the stored mean cascade on this condition is 129 tokens (Table~\ref{tab:cascade}). Unique-word and $n$-gram anchors can jump to the matching paragraph anywhere on the page.

\begin{figure*}[t]
\centering
\includegraphics[width=\linewidth]{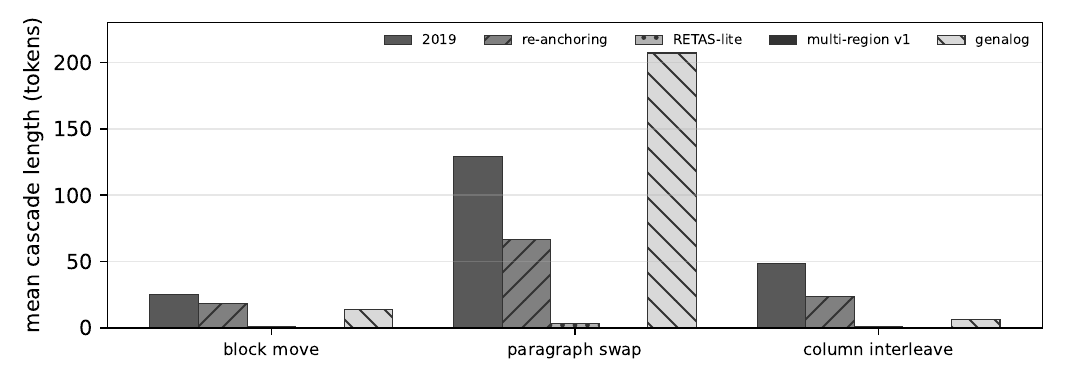}
\caption{Mean cascade length on the synthetic ablation at medium severity and 6\% character noise, the same stored rows that fill Table~\ref{tab:cascade}. No new aligner runs.}\label{fig:cascadezoning}
\end{figure*}

Figure~\ref{fig:cascadezoning} plots those same stored rows. On paragraph swap the mean cascade is 129 tokens for the 2019 method, 66.5 after re-anchoring, 207 for genalog, 3.3 for \lite and 0 for \mr. On column interleave the same order is 48.8, 23.6, 6.3, 1.3 and 0.1; on a single block move it is 25.3, 18.2, 13.7, 0.7 and 0.0. The figure is a companion to Table~\ref{tab:cascade}, not a new experiment.

\subsection{Real reading order without OCR noise}\label{sec:res-ro}

\begin{table*}[t]
\caption{KB IMPACT newspaper pages whose geometric order differs from the stored XML-traversal linearization of PAGE region references (40 of 50 scanned pages). The ground-truth text is used on both sides, so there is no OCR noise; only the reading order differs. F1 with 95\% bootstrap CI over pages, recovery $R_{10}$ and runtime per page.}\label{tab:impactro}
\centering\small
\begin{tabular}{@{}lccc@{}}
\toprule
Aligner & F1 & $R_{10}$ & Runtime (s) \\
\midrule
Location-aware (2019) & 0.573 [0.435, 0.707] & 0.550 & 0.008 \\
+ re-anchoring & 0.578 [0.441, 0.712] & 0.552 & 0.031 \\
RETAS-lite & 0.986 [0.972, 0.995] & 0.987 & 0.005 \\
Multi-region v1 & 0.987 [0.969, 0.999] & 0.995 & 0.009 \\
genalog (RETAS) & 0.668 [0.549, 0.792] & 0.643 & 2.544 \\
Java RETAS & 0.590 [0.557, 0.623] & 0.870 & 0.102  \\
\botrule
\end{tabular}
\end{table*}

Table~\ref{tab:impactro} lists the stored Step-3d reading-order scores: \mr 0.987, \lite 0.986, genalog 0.668, the 2019 method 0.573 and re-anchoring 0.578. Java RETAS was rerun on the same Step-3d page list, giving F1 0.590 $[0.557, 0.623]$ and recovery $R_{10}=0.870$. A different-draw Java-harness run is excluded from cross-method comparisons.

\subsection{Real OCR with geometry gold}\label{sec:res-geo}

\begin{table*}[t]
\caption{Real Tesseract OCR on 200 held-out KB IMPACT pages (seed 42; at most 600 OCR words per page), scored against geometry-derived gold. ``All'' scores every OCR word that received a gold match; ``high-conf.'' keeps only words whose gold match is exact or has NED $\le 0.15$ with IoU $\ge 0.05$. The gold is approximate (Sect.~\ref{sec:geogold}), with the limitations described there. Mean F1 with 95\% bootstrap CI over pages.}\label{tab:impactgeo}
\centering\small
\begin{tabular}{@{}lcc@{}}
\toprule
Aligner & F1 (all gold) & F1 (high-conf. gold) \\
\midrule
Multi-region v1 & 0.432 [0.404, 0.460] & 0.650 [0.623, 0.675]  \\
Multi-region v2 & 0.403 [0.375, 0.432] & 0.607 [0.579, 0.634]  \\
RETAS-lite & 0.354 [0.326, 0.381] & 0.504 [0.477, 0.530]  \\
genalog (RETAS) & 0.254 [0.230, 0.277] & 0.371 [0.342, 0.399]  \\
Java RETAS & 0.176 [0.164, 0.187] & 0.273 [0.260, 0.284]  \\
+ re-anchoring & 0.042 [0.034, 0.051] & 0.059 [0.050, 0.071]  \\
Location-aware (2019) & 0.038 [0.031, 0.047] & 0.053 [0.044, 0.064]  \\
\botrule
\end{tabular}
\end{table*}

\begin{figure}[t]
\centering
\includegraphics[width=0.78\linewidth]{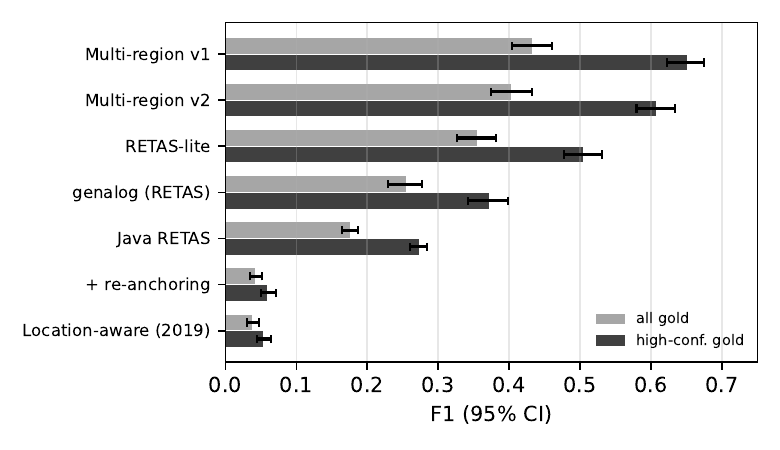}
\caption{F1 on the 200 KB IMPACT pages with Tesseract OCR (Table~\ref{tab:impactgeo}), on all and high-confidence geometry gold, with bootstrap intervals.}\label{fig:impactgeo}
\end{figure}

On the high-confidence gold, stored F1 is \mr v1 0.650, \lite 0.504, genalog 0.371, Java RETAS 0.273 and the 2019 method 0.053. The stored all-gold differences are $+0.078$ $[+0.068, +0.087]$ for \mr minus \lite, $+0.256$ $[+0.235, +0.277]$ for \mr minus Java RETAS, and $+0.178$ $[+0.155, +0.201]$ for \mr minus genalog. The first two were recomputed from stored per-page scores. The geometry-derived gold remains approximate.

\begin{figure*}[t]
\centering
\includegraphics[width=\linewidth]{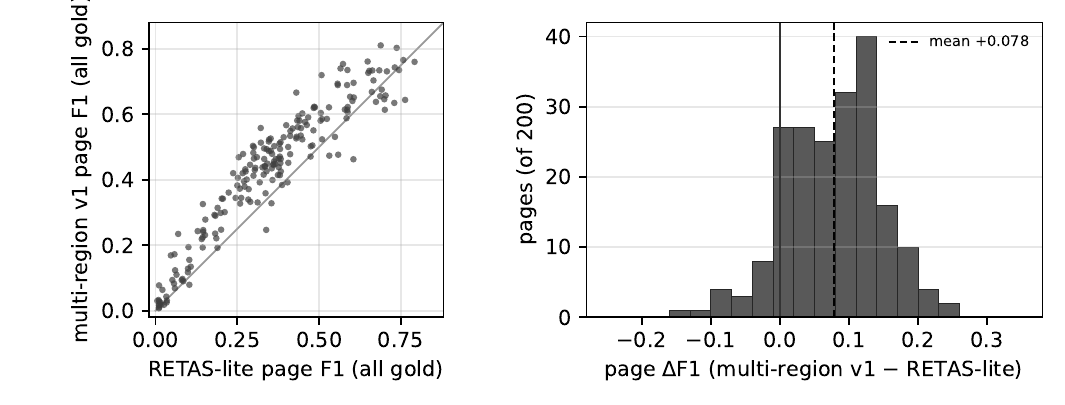}
\caption{Page-level all-gold F1 on the 200 KB IMPACT pages with Tesseract OCR. Left: \mr v1 against \lite (the diagonal is equality). Right: histogram of the per-page difference, with the mean marked. Recomputed from the stored per-page scores that underlie Table~\ref{tab:impactgeo}.}\label{fig:pagef1}
\end{figure*}

Those means hide a skewed page-level picture (Fig.~\ref{fig:pagef1}). Recomputed from the stored all-gold F1 of the same 200 pages, \mr has page-mean 0.432 and \lite 0.354 (difference $+0.078$), matching the stored paired gap; the 2019 method has page-mean 0.038 and falls below 0.10 on 191 of the 200 pages. \mr exceeds \lite on 175 pages and is behind on 25, with no ties. The median page difference is $+0.085$. Of the 175 pages where \mr is ahead, the 52 largest positive gaps account for at least half of the summed positive gap, and the 99 largest account for at least 80\%: the headline $+0.078$ is not a uniform lift. \mr reaches F1 of at least 0.50 on 80 pages, and on 75 of those the 2019 method is below 0.15.

Before constructing geometry gold, the same pages were scored for agreement with genalog: stored values are \mr 0.185, \lite 0.172, Java RETAS 0.116 and the 2019 method 0.021, with \mr minus \lite $+0.012$ $[+0.008, +0.017]$. These are agreement records, not correctness measurements.

\subsection{Pages with human word boxes}\label{sec:res-wordgold}

\begin{table*}[t]
\caption{Fourteen CC0 pages with human word boxes (blackletter and antiqua prints), Tesseract OCR, gold from word-box IoU and NED. Small sample; the gold-matching and domain limitations still apply.}\label{tab:wordgold}
\centering\small
\begin{tabular}{@{}lcc@{}}
\toprule
Aligner & F1 (all gold) & F1 (high-conf. gold) \\
\midrule
Multi-region v1 & 0.742 [0.687, 0.795] & 0.948 [0.918, 0.975] \\
Multi-region v2 & 0.735 [0.677, 0.791] & 0.947 [0.917, 0.973] \\
RETAS-lite & 0.698 [0.629, 0.765] & 0.910 [0.869, 0.949] \\
Location-aware (2019) & 0.483 [0.373, 0.592] & 0.694 [0.547, 0.835] \\
genalog (RETAS) & 0.492 [0.341, 0.628] & 0.647 [0.437, 0.833] \\
Java RETAS & 0.378 [0.323, 0.436] & 0.663 [0.609, 0.712] \\
\botrule
\end{tabular}
\end{table*}

Table~\ref{tab:wordgold} reports stored 14-page word-box scores: \mr v1 F1 0.742 (all gold) and 0.948 (high-confidence), and \lite 0.698 and 0.910; the stored all-gold difference is $+0.043$ $[+0.024, +0.065]$. Stored all-gold F1 for the 2019 method, genalog and Java RETAS is 0.483, 0.492 and 0.378. These early modern prints are not a multi-column newspaper evaluation.

\subsection{Stress tests}\label{sec:res-stress}

\begin{table*}[t]
\caption{Stress tests with frozen parameters, 10 seeds. Noise rows use column interleave with the given per-word character-noise rate; region rows split the page into $k$ regions; short pages have about 96 words; the boilerplate page has 693 words containing 59 distinct word types. $\Delta$ is the paired bootstrap mean of multi-region v1 minus RETAS-lite with 95\% CI, and $p$ twice the share of bootstrap means on the other side of zero, capped at one (``$<$0.001'' means none of the resamples).}\label{tab:stress}
\centering\footnotesize\setlength{\tabcolsep}{4pt}
\begin{tabular}{@{}lcccccc@{}}
\toprule
Condition & MR v1 & RETAS-lite & Java RETAS & 2019 & $\Delta$ (MR v1 $-$ lite) & $p$ \\
\midrule
Noise 0.10 & 0.972 & 0.893 & 0.476 & 0.264 & +0.079 [+0.076, +0.083] & $<$0.001 \\
Noise 0.15 & 0.973 & 0.889 & 0.457 & 0.243 & +0.084 [+0.078, +0.091] & $<$0.001 \\
Noise 0.20 & 0.957 & 0.873 & 0.436 & 0.228 & +0.084 [+0.077, +0.094] & $<$0.001 \\
Noise 0.25 & 0.962 & 0.870 & 0.416 & 0.268 & +0.093 [+0.086, +0.099] & $<$0.001 \\
\midrule
8 regions & 0.998 & 0.981 & 0.432 & 0.141 & +0.017 [+0.014, +0.021] & $<$0.001 \\
15 regions & 0.995 & 0.969 & 0.385 & 0.101 & +0.027 [+0.022, +0.031] & $<$0.001 \\
30 regions & 0.989 & 0.942 & 0.382 & 0.131 & +0.046 [+0.039, +0.054] & $<$0.001 \\
\midrule
Short pages & 1.000 & 0.998 & 0.245 & 0.321 & +0.002 [+0.000, +0.004] & 0.23 \\
Boilerplate & 0.125 & 0.030 & 0.069 & 0.456 & +0.094 [+0.087, +0.104] & $<$0.001 \\
\botrule
\end{tabular}
\end{table*}

\begin{figure}[t]
\centering
\includegraphics[width=\linewidth]{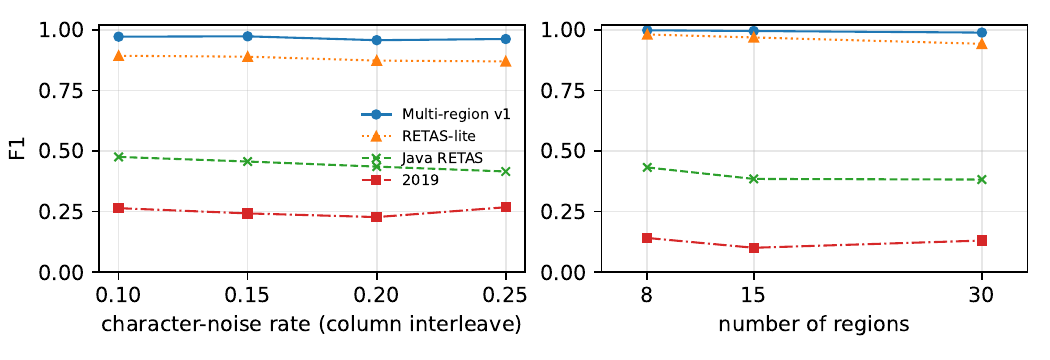}
\caption{Stress tests of Table~\ref{tab:stress}: F1 against the character-noise rate on column interleave (left) and against the number of text regions (right). Means over 10 seeds.}\label{fig:stress}
\end{figure}

Table~\ref{tab:stress} and Fig.~\ref{fig:stress} vary one difficulty at a time with the v1 parameters frozen. Under character noise from 10\% to 25\%, \mr stays between 0.957 and 0.973 while \lite drops to 0.870--0.893; the paired gap is $+0.079$ to $+0.093$. Noise destroys some unique words and creates false ones; we think the $n$-gram anchors and chunk overlaps of \mr are less sensitive to a single corrupted word than unique words alone. When the page is split into more regions, the gap grows from $+0.017$ at 8 regions to $+0.046$ at 30. On short pages of about 96 words both methods are at the ceiling and the difference is not significant ($p = 0.23$). In the noise and region conditions Java RETAS stays between 0.38 and 0.48 and the 2019 method between 0.10 and 0.27.

The boilerplate page is different. With 59 distinct word types in 693 tokens, every anchor-based method collapses: \mr 0.125, Java RETAS 0.069 and \lite 0.030. The 2019 location-aware method scores 0.456, more than three times the best anchor-based method. On this page, knowing roughly where one is in the ground truth is worth more than any anchor. It is a single synthetic text, so we do not know how far the result generalises, but we do not think it is an artefact of the generator, and it is the reason v2 tries to fall back on the 2019 method when anchors are scarce.

\subsection{Long documents and the second version}\label{sec:res-long}

\begin{table*}[t]
\caption{Long documents built from distinct text (no repetition), column interleave, medium severity, 6\% noise. Mean F1 over seeds (number of seeds in parentheses; range over seeds in Fig.~\ref{fig:longdoc}) and mean runtime per document in seconds. v1 and RETAS-lite are from the Step-7 scaling study; v2 is from the separate Step-8 re-run. At 8{,}000 tokens and above, v2 hands the document to RETAS-lite, so its F1 equals RETAS-lite by construction.}\label{tab:longdoc}
\centering\footnotesize\setlength{\tabcolsep}{4pt}
\begin{tabular}{@{}lcccccc@{}}
\toprule
 & \multicolumn{2}{c}{Multi-region v1} & \multicolumn{2}{c}{Multi-region v2} & \multicolumn{2}{c}{RETAS-lite} \\
\cmidrule(lr){2-3}\cmidrule(lr){4-5}\cmidrule(lr){6-7}
Tokens (seeds) & F1 & s & F1 & s & F1 & s \\
\midrule
1{,}000 (5) & 0.771 & 0.021 & 0.768 & 0.015 & 0.504 & 0.003 \\
5{,}000 (5) & 0.423 & 0.588 & 0.418 & 0.219 & 0.294 & 0.011 \\
10{,}000 (5) & 0.334 & 2.533 & 0.326 & 0.036 & 0.326 & 0.036 \\
25{,}000 (3) & 0.284 & 14.7 & 0.343 & 0.153 & 0.343 & 0.148 \\
50{,}000 (3) & 0.256 & 61.5 & 0.341 & 0.511 & 0.341 & 0.507 \\
\botrule
\end{tabular}
\end{table*}

\begin{figure}[t]
\centering
\includegraphics[width=\linewidth]{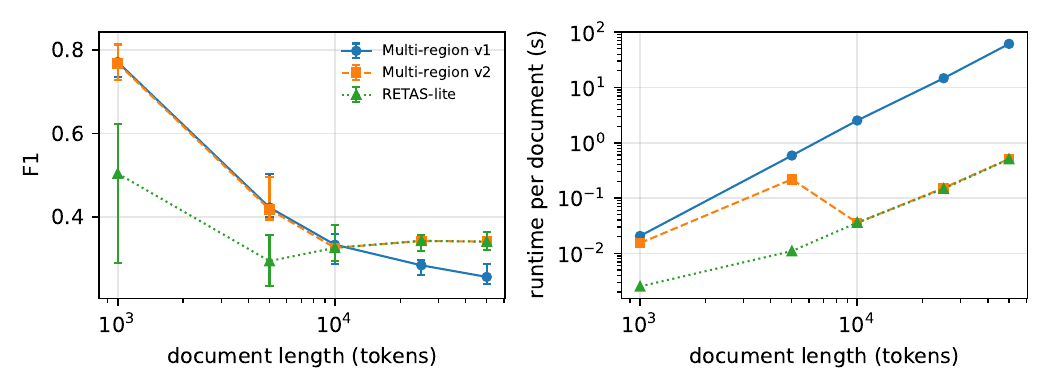}
\caption{Long documents of distinct text (Table~\ref{tab:longdoc}): mean F1 with the range over seeds (left) and runtime per document on a log scale (right).}\label{fig:longdoc}
\end{figure}

The \mr aligner was designed for pages, and Table~\ref{tab:longdoc} and Fig.~\ref{fig:longdoc} show what happens beyond them. All methods score lower at 50{,}000 tokens than at 1{,}000. We think this is because the same zoning pattern over a longer text creates more breaks and leaves a smaller share of words unique. Up to 5{,}000 tokens \mr v1 is clearly ahead of \lite (0.771 against 0.504 at 1{,}000 tokens, 0.423 against 0.294 at 5{,}000). At 10{,}000 they are level, and at 25{,}000 and 50{,}000 \lite is better (0.341 against 0.256 at 50{,}000). The runtime of v1 grows from 0.021\,s at 1{,}000 tokens to 2.53\,s at 10{,}000 and 61.5\,s at 50{,}000, roughly quadratic, while \lite stays under a second. Profiling attributes 4.46 of 5.29\,s at about 10{,}000 tokens to the chunk-to-window scoring of step~3. An earlier long-document test built its documents by repeating the base text; there both methods collapsed (\mr 0.424 at 1{,}000 tokens and 0.016 at 50{,}000, \lite 0.026 and 0.001) because repetition removes uniqueness, so that test measures repetition rather than length and we do not use it further.

\begin{table*}[t]
\caption{Multi-region v1 and v2, both frozen, on the held-out evaluations (Step-8 re-runs unless noted). $\Delta$ is the paired bootstrap mean of v2 minus v1 with 95\% CI. $^{a}$Three seeds only, so we give the mean and the range over seeds instead of a bootstrap interval. Synthetic and long-document entries are distinct from the newspaper and word-box evaluations. The TREC row uses synthetic character noise and column interleaving, not real OCR.}\label{tab:v2}
\centering\footnotesize\setlength{\tabcolsep}{4pt}
\begin{tabular}{@{}lccc@{}}
\toprule
Evaluation & v1 F1 & v2 F1 & $\Delta$ (v2 $-$ v1) \\
\midrule
IMPACT test-200, all gold & 0.432 [0.404, 0.460] & 0.403 [0.375, 0.432] & $-$0.029 [$-$0.032, $-$0.026] \\
IMPACT test-200, high-conf. & 0.650 [0.623, 0.675] & 0.607 [0.579, 0.634] & $-$0.043 [$-$0.048, $-$0.038] \\
Word-box gold, all & 0.742 [0.687, 0.795] & 0.735 [0.677, 0.791] & $-$0.007 [$-$0.012, $-$0.002] \\
TREC synthetic (Step 8) & 0.945 [0.935, 0.955] & 0.942 [0.933, 0.952] & $-$0.002 [$-$0.004, $-$0.001] \\
Boilerplate stress & 0.125 [0.118, 0.134] & 0.159 [0.151, 0.168] & +0.034 [+0.029, +0.041] \\
Long document, 50k tokens$^{a}$ & 0.256 (0.239 to 0.287) & 0.341 (0.320 to 0.363) & +0.085 (+0.077 to +0.096) \\
\botrule
\end{tabular}
\end{table*}

The frozen v2 bypasses v1's quadratic chunk-to-window scoring on long documents. In the separate Step-8 side-by-side re-run on the same machine, v1 took 57.7\,s at 50{,}000 tokens rather than the scaling study's 61.5\,s; this is run-to-run timing noise, not a change to v1. With its fallback, v2 is identical to \lite from 8{,}000 tokens on and runs in 0.51\,s at 50{,}000 tokens instead of 57.7\,s, with F1 0.341 instead of 0.256 (Table~\ref{tab:v2}). Below the threshold it is less than 0.01 below v1, and its indexed chunk assignment is faster (0.22\,s against 0.59\,s at 5{,}000 tokens).

In the IMPACT comparison in Table~\ref{tab:v2}, high-confidence F1 is v2 0.607 versus v1 0.650, paired difference $-0.043$. Recorded fallback counts are scarcity 48/200 and long-document 0/200 under the 600-token cap. These counts do not establish the cause of a score difference. The word-box all-gold difference is stored as $-0.007$. In the synthetic boilerplate test, v2 gains 0.034 over v1 (0.159 versus 0.125), below the pre-specified 0.15 target and the 2019 method's 0.456. The scarcity merge applies the 2019 prior only to unmatched spans. We withhold an explanation of its residual errors pending independent annotation.

\subsection{TREC-5: a reproduction attempt}\label{sec:res-trec}

\begin{table*}[t]
\caption{TREC-5 Confusion Track, 30 randomly sampled Federal Register documents. There is no word-level gold, so we report agreement F1 with genalog's alignment as a soft reference (genalog scores 1 by definition) and the fraction of OCR words that receive any match (``matched''), the closest analogue we can compute to the 2019 success rate. Neither is a correctness score.}\label{tab:trec}
\centering\footnotesize\setlength{\tabcolsep}{4pt}
\begin{tabular}{@{}lcccc@{}}
\toprule
 & \multicolumn{2}{c}{degrade5 ($\approx$5\% CER)} & \multicolumn{2}{c}{degrade20 ($\approx$20\% CER)} \\
\cmidrule(lr){2-3}\cmidrule(lr){4-5}
Aligner & Agreement F1 & Matched & Agreement F1 & Matched \\
\midrule
Location-aware (2019) & 0.888 [0.827, 0.933] & 0.957 & 0.102 [0.076, 0.131] & 0.740 \\
+ re-anchoring & 0.893 [0.834, 0.936] & 0.959 & 0.107 [0.081, 0.137] & 0.743 \\
RETAS-lite & 0.934 [0.921, 0.945] & 0.912 & 0.378 [0.354, 0.404] & 0.375 \\
Multi-region v1 & 0.944 [0.934, 0.954] & 0.933 & 0.541 [0.521, 0.561] & 0.671 \\
Java RETAS & 0.450 [0.438, 0.463] & 0.311 & 0.179 [0.170, 0.188] & 0.145 \\
genalog (RETAS) (reference) & 1.000 [1.000, 1.000] & 1.000 & 1.000 [1.000, 1.000] & 1.000 \\
\botrule
\end{tabular}
\end{table*}

Table~\ref{tab:trec} returns to the data of the 2019 paper. We cannot rebuild its split: TREC-5 does not mark which documents have zoning problems, and the 2019 document lists and scoring script are not part of the materials we had. On 30 random documents of degrade5, our reimplementation of the 2019 method matches 95.7\% of OCR words, between the 98.5\% and 81\% reported in 2019 for the two groups; we cannot say more than that.

The table also shows why the matched rate is not a correctness score. On degrade20 the 2019 method still matches 74.0\% of words, but its agreement with genalog's alignment is only 0.102. Against genalog as the reference, \mr agrees slightly more than \lite on degrade5 (0.944 against 0.934, paired $+0.010$ $[+0.006, +0.015]$) and much more on degrade20 (0.541 against 0.378, $+0.163$ $[+0.145, +0.183]$). Java RETAS agrees with genalog at 0.450 and 0.179, although both implement the same published scheme. Because genalog is the reference here, these numbers measure agreement with one aligner, not correctness.

With synthetic character noise and column interleaving on TREC originals (20 documents, medium column interleave, 10\% noise), the canonical Step-4 comparison gives \mr v1 F1 0.914 $[0.880, 0.939]$, \lite 0.704 $[0.667, 0.741]$ and Java RETAS 0.304 $[0.280, 0.327]$. These are correctness scores against the synthetic generator's word-level gold, not evidence on real OCR and not agreement with genalog. A separate Step-8 synthetic-noise and column-interleaving re-run, used only for the v1--v2 comparison in Table~\ref{tab:v2}, gives v1 0.945, v2 0.942 and \lite 0.736. The two runs use different draws and seeds and must not be pooled.

\section{Discussion and limitations}\label{sec:discussion}

\paragraph{What changed since 2019.} The 2019 paper treated zoning as a failure of a tracker that loses its place, and proposed a location-aware tracker that searches a window around its last match. Our re-anchoring extension repairs short drifts but does not resolve large reading-order changes in these tests, because its search starts from the last good match. The methods that handle large displacements start from the other end. They first match what can be matched with high confidence anywhere on the page, and only then fill in the rest.

\paragraph{When to use which aligner.} The synthetic tests suggest that \mr v1 is suited to page-sized inputs. The real-OCR comparison is subject to the approximate-gold limitations below. On the tested long synthetic documents, \lite and v2 (which falls back to it) had higher F1 and lower runtime than v1 at the largest lengths. On our synthetic boilerplate page with few distinct word types, the 2019 location-aware method had the highest F1 among the tested methods. We have not established whether this result holds for forms or repeated notices.

\paragraph{Limitations.} The main real-OCR test uses gold derived from region boxes, with an estimated error in about a quarter of the matched pairs, and the only test with true word boxes has 14 pages. The gold and the article-order convention are not independently validated. All OCR came from one engine, Tesseract~5. Our RETAS baseline is a simplified reimplementation, and our genalog and Java RETAS results depend on our conversion of their output to word mappings, which we checked for Java RETAS only. The real-OCR data is historical: 17th- to 19th-century Dutch newspapers and 16th- and 17th-century prints. We found no open data set of modern multi-column pages with word-level ground truth. The 2019 TREC-5 split could not be reproduced. The \mr aligner is quadratic in document length, and the v2 attempt to combine it with the 2019 prior did not reach its pre-specified target. The synthetic boilerplate fixture does not establish that a location prior will help on real anchor-scarce pages. We have not independently annotated the remaining disagreements, so we do not classify their causes or propose a new aligner on that basis. In a small LoRA corrector pilot under a region-level evaluation protocol, alignment-derived training pairs did not improve CER over raw OCR; whether better alignment helps downstream correction remains open.

\paragraph{Reproducibility.} All aligners are deterministic and run on CPU. Parameters are frozen and listed in Sect.~\ref{sec:methods}. The data sets are obtainable: TREC-5 from NIST, the \impact ground truth under CC0, and the word-box corpora under CC0. The alignment code will be released under the Apache License~2.0 at \url{https://github.com/sniperjf/zone-robust-ocr-align} at submission.

\section{Conclusion}\label{sec:conclusion}

We revisited the Zoning Problem from our 2019 alignment paper with OCR-side permutations and word-level construction gold. The multi-region aligner combines unique-word and n-gram anchors, local matching and chunk-to-window assignment; the synthetic experiments examine where local tracking and anchor-based matching fail. Re-anchoring gave a modest gain on the synthetic permutations; on the 200 real pages it was close to the 2019 baseline. The multi-region v1 implementation has a quadratic scaling cost. Approximate gold and the article-order convention remain limitations of the intrinsic evaluations. Independent residual-error annotation and modern multi-column word-level ground truth remain necessary.

\backmatter
\bmhead{Acknowledgements}

We thank the Chambers-Grundy Center for Transformative Neuroscience and the Department of Brain Health at the University of Nevada, Las Vegas, for access to an NVIDIA H200 GPU used in a small LoRA corrector pilot (1.691 GPU-hours of device time), mentioned only as a limitation in Sect.~\ref{sec:discussion} and not tabulated as a result. Alignment experiments were run on an AMD EPYC 9175F CPU using Python~3.12, as reported in Sect.~\ref{sec:setup}.

\section*{Statements and Declarations}

\paragraph{Funding.} No funding was received for conducting this study. Compute resources are acknowledged above.

\paragraph{Competing interests.} The author has no competing interests to declare that are relevant to the content of this article.

\paragraph{Ethics approval and consent.} Not applicable.

\paragraph{Data availability.} The TREC-5 Confusion Track data is available from NIST (\url{https://trec.nist.gov/data/t5_confusion.html}). The KB IMPACT ground truth is available from the KB under CC0 (\url{https://lab.kb.nl/dataset/ground-truth-impact-project}). The word-box corpora (17\_frak\_simple, 17\_frak\_complex and 16\_ant\_simple) are available under CC0 from \url{https://github.com/tboenig/17_frak_simple}, \url{https://github.com/tboenig/17_frak_complex} and \url{https://github.com/tboenig/16_ant_simple}. We do not redistribute any of these data sets. Derived geometry gold and page lists will be released with the code if rights permit; they are held pending a rights check. The alignment code will be released under the Apache License~2.0 at \url{https://github.com/sniperjf/zone-robust-ocr-align} at submission.

\paragraph{Code availability.} The alignment code will be released under the Apache License~2.0 at \url{https://github.com/sniperjf/zone-robust-ocr-align} at submission.

\paragraph{Author contributions.} J.R.F.C. (ORCID 0000-0001-7917-5169) is the sole author and was responsible for conceptualization, methods, software, analysis and writing.

\paragraph{Declaration of AI-assisted writing and tools.} During the preparation of this work, the author used AI assistants, namely Grok Bot (xAI), Claude via Claude Code (Anthropic), and the OpenAI Codex CLI, to help with language refinement, manuscript organization, drafting portions of the text from author-provided research materials, and implementing and running experiment code under the author's direction. The OpenAI Codex CLI was also used to proofread manuscript wording against source files. The author designed the study, reviewed and edited the content as needed, and takes full responsibility for the content of this manuscript.

\begin{biography}{\biofigure{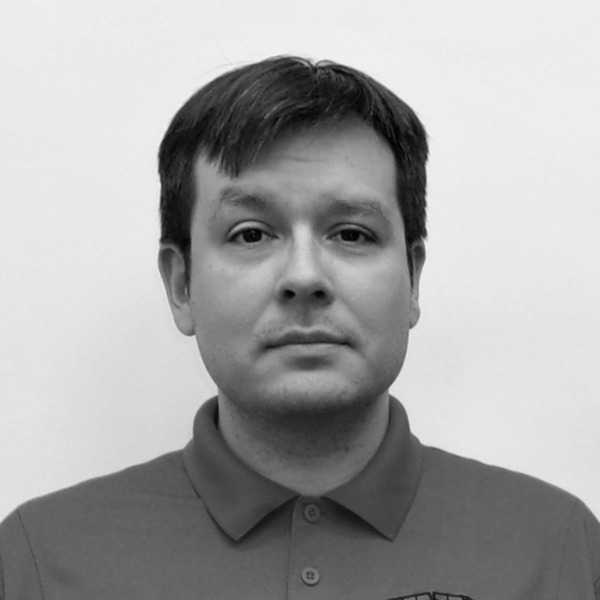}}{
Jorge Ram\'on Fonseca Cacho is an Assistant Professor (tenure track) of Computer Science at the University of Nevada, Las Vegas (UNLV), where he has taught since 2018. He received his Ph.D.\ in Computer Science from UNLV in 2019 with a dissertation on improving OCR post-processing with machine learning, and a B.A.\ in Film and Theater from UNLV. His research covers OCR post-processing, NLP, reproducible research, and trustworthy, verifiable LLM systems that trace outputs to their sources. As Co-Investigator on Alzheimer's clinical-trial observatory projects, he leads AI and data engineering for large-scale biomedical text extraction.
}
\end{biography}

\bibliography{refs}

\end{document}